\pdfoutput=1
\documentclass[11pt]{article}
\usepackage[]{acl}
\usepackage{times}
\usepackage{latexsym}
\usepackage{lettrine}
\usepackage[T1]{fontenc}
\usepackage[utf8]{inputenc}
\usepackage{microtype}
\usepackage{url}
\usepackage{multicol}
\usepackage{multirow}
\usepackage{graphicx}
\usepackage{subcaption}
\usepackage{placeins}
\usepackage{makecell}
\usepackage{amsmath}
\usepackage{amsfonts}
\usepackage{subcaption}
\usepackage{float}
\usepackage{titlesec}
\usepackage{tikz}
\usetikzlibrary{pgfplots.groupplots}
\usepackage{pgfplots}\pgfplotsset{compat=1.17} 
\usepackage[font=small]{caption}

\title{A Unified Model for Reverse Dictionary and Definition Modelling}

\author{Pinzhen Chen \qquad\qquad Zheng Zhao \\
  School of Informatics, University of Edinburgh \\
  \texttt{\{pinzhen.chen,zheng.zhao\}@ed.ac.uk}}

\begin{document}
\maketitle
\begin{abstract}
We build a dual-way neural dictionary to retrieve words given definitions, and produce definitions for queried words. The model learns the two tasks simultaneously and handles unknown words via embeddings. It casts a word or a definition to the same representation space through a shared layer, then generates the other form in a multi-task fashion. Our method achieves promising automatic scores on previous benchmarks without extra resources. Human annotators prefer the model's outputs in both reference-less and reference-based evaluation, indicating its practicality. Analysis suggests that multiple objectives benefit learning. 

\end{abstract}

\section{Introduction}
A monolingual dictionary is a large-scale collection of words paired with their definitions. The main use of such a resource is to find a word or a definition having known the other. Formally, the task of generating a textual definition from a word is named \textit{definition modelling}; the inverse task of retrieving a word given a definition is called \textit{reverse dictionary}. Lately, the two tasks are approached using neural networks \citep{hill-etal-2016-learning-understand, noraset-etal-2017-definition}, and in turn they help researchers better understand word sense and embeddings. Research can further benefit low-resource languages where high-quality dictionaries are not available \citep{yan-etal-2020-bert}. Finally, practical applications include language education, writing assistance, semantic search, etc.

While previous works solve one problem at a time, we argue that both tasks can be learned and dealt with concurrently, based on the intuition that a word and its definition share the same meaning. We design a neural model to embed words and definitions into a shared semantic space, and generate them from this space. Consequently, the training paradigm can include reconstruction and embedding similarity tasks. Such a system can be viewed as a neural dictionary that supports two-way indexing and querying. In our experiments, jointly learning both tasks does not increase the total model size, yet demonstrates ease and effectiveness. Our code is publicly available.\footnote{\url{https://github.com/PinzhenChen/unifiedRevdicDefmod}}

\section{Related Work}
Although research on the two tasks can be traced back to the early 2000s, recent research has shifted towards neural networks, which we describe here. 

\paragraph{Reverse dictionary} \citet{hill-etal-2016-learning-understand} pioneer the use of RNN and bag-of-words models to convert texts to word vectors, on top of which \citet{morinaga-yamaguchi-2018-improvement} add an extra word category classifier. \citet{pilehvar-2019-importance} integrates super-sense into target embeddings to disambiguate polysemous words. \citet{zhang-etal-2020-multi-channel} design a multi-channel network to predict a word with its features like category, POS tag, morpheme, sememe, etc.

Nonetheless, our work tackles the problem without using linguistically annotated resources. The proposed framework learns autoencodings for definitions and words, instead of mapping texts to plain word vectors. From this aspect, \citet{bosc-vincent-2018-auto} train word embeddings via definition reconstruction.

\paragraph{Definition modelling}\label{sec:2-defmod} \citet{noraset-etal-2017-definition} use RNNs for definition generation, followed by \citet{gadetsky-etal-2018-conditional} who add attention and word context, as well as \citet{chang-etal-2018-xsense} whose model projects words and contexts to a sparse space, then generates from selected dimensions only. \citet{mickus-etal-2019-mark}'s model encodes a context sentence and marks the word of interest, whereas \citet{bevilacqua-etal-2020-generationary}'s defines a flexible span of words. Apart from generating definitions freely, \citet{chang-chen-2019-word} take a new perspective of re-formulating the generation task to definition retrieval from a dictionary.

\section{Methodology}
\subsection{A unified model with multi-task training}
A word (embedding) and its definition share the same meaning, even though they exist in different surface forms. When we model their semantics using a neural method, we hypothesize that a word and its definition can be encoded into a consistent representation space. This gives rise to our core architecture in the paper: a model that transforms inputs into a shared embedding space that can represent both words and definitions. We then have downstream modules that convert the shared embeddings back to words or definitions. Essentially, the shared representation can be viewed as an autoencoding of the meaning of a word and its definition. In the learning process, definition modelling and reverse dictionary are jointly trained to aid each other; yet at inference time, only half of the network needs to be used to perform either task.

\begin{figure}[htb]
    \includegraphics[width=1.0\linewidth]{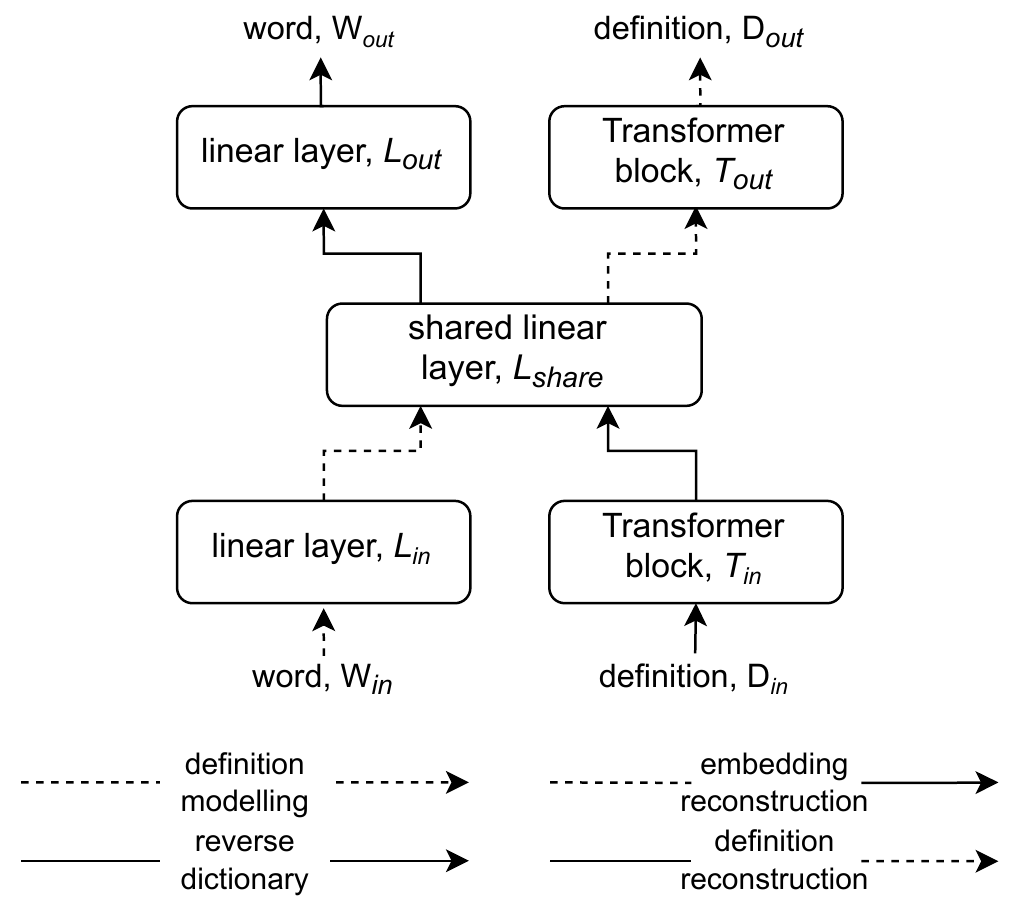}
    \caption{An illustration of our designed model.}
    \label{fig:arch1}
\end{figure}

The proposed architecture with four sub-task workflows is illustrated in Figure~\ref{fig:arch1}. The autoencoding capability is accomplished through a shared linear layer \textit{L}$_{share}$ between the encoder and the decoder networks, the output of which is the encoded words and definitions. We use linear layers \textit{L}$_{in}$ and \textit{L}$_{out}$ to process words W$_{in}$ and W$_{out}$ before and after the shared layer. Likewise, we have definitions D$_{in}$ and D$_{out}$ converted to and from the shared layer, using Transformer blocks \textit{T}$_{in}$ and \textit{T}$_{out}$ \citep{vaswani-etal-2017-attention}. In addition, we encourage the shared layer's representations of the input word W$_{in}$ and definition D$_{in}$ to be as close as possible. The Transformer blocks operate on self-attention but not encoder-decoder attention, i.e. Transformer blocks do not attend to each other, so as to force all information to flow through the autoencoding bottleneck.

With an embedding distance $\mathrm{embed\_dist()}$ and a token-level loss $\mathrm{token\_loss()}$, canonical reverse dictionary and definition modelling have losses:

{\small\setlength{\abovedisplayskip}{0ex}
    \begin{align}
        \mathcal{L}_\mathrm{revdic}=&\mathrm{embed\_dist}(\mathrm{W}_\mathit{gold}, \mathit{L}_\mathit{out}(\mathit{L}_\mathit{share}(\mathit{T}_\mathit{in}(\mathrm{D}_\mathit{in})))) \nonumber \\
        \mathcal{L}_\mathrm{defmod}=&\mathrm{token\_loss}(\mathrm{D}_\mathit{gold},  \mathit{T}_\mathit{out}(\mathit{L}_\mathit{share}(\mathit{L}_\mathit{in}(\mathrm{W}_\mathit{in})))) \nonumber
    \end{align}
}%

\noindent Our model also optimizes on the losses from word and definition reconstruction (autoencoding):

{\small\setlength{\abovedisplayskip}{0ex}
    \begin{align}
        \mathcal{L}_\mathrm{wordAE}=&\mathrm{embed\_dist}(\mathrm{W}_\mathit{gold}, \mathit{L}_\mathit{out}(\mathit{L}_\mathit{share}(\mathit{L}_\mathit{in}(\mathrm{W}_\mathit{in})))) \nonumber \\
        \mathcal{L}_\mathrm{defAE}=&\mathrm{token\_loss}(\mathrm{D}_\mathit{gold}, \mathit{T}_\mathit{out}(\mathit{L}_\mathit{share}(\mathit{T}_\mathit{in}(\mathrm{D}_\mathit{in})))) \nonumber
    \end{align}
}%

\noindent The distance between a pair of word and definition representations from the shared layer is:

{\small\setlength{\abovedisplayskip}{0ex}
    \begin{align}
        \mathcal{L}_\mathrm{sim}=\mathrm{embed\_dist}(\mathit{L}_\mathit{share}(\mathit{T}_\mathit{in}(\mathrm{D}_\mathit{in})), \mathit{L}_\mathit{share}(\mathit{L}_\mathit{in}(\mathrm{W}_\mathit{in}))) \nonumber
    \end{align}
}%

\noindent Finally, our training minimizes the overall loss $\mathcal{L}$ that adds all above losses weighted equally:

{\small\setlength{\abovedisplayskip}{0ex}
    \begin{align}
        \mathcal{L} = \mathcal{L}_\mathrm{revdic} + \mathcal{L}_\mathrm{defmod} + \mathcal{L}_\mathrm{wordAE} + \mathcal{L}_\mathrm{defAE} + \mathcal{L}_\mathrm{sim} \nonumber
    \end{align}
}%

\subsection{Word-sense disambiguation}
A word is often associated with multiple definitions due to the presence of polysemy, sense granularity, etc. In our practice, the one-to-many word-definition relationship does not harm reverse dictionary, since our model can master mapping different definitions into the same word vector. However, it is problematic for definition modelling, as telling the exact word sense without context is hard. Thus, we embed words in their usage context (supplied in the data we use) using BERT \citep{devlin-etal-2019-bert}. We sum up the sub-word embeddings for each word if it is segmented by BERT.

\section{Experiments and Results}
\label{sec:4-exp}

\subsection{Data and evaluation}
\textbf{\textsc{Hill}}: we evaluate reverse dictionary on \citet{hill-etal-2016-learning-understand}'s English data. There are roughly 100k words and 900k word-definition pairs. Three test sets are present to test a system's memorizing and generalizing capabilities: 500 \textit{seen} from training data, 500 \textit{unseen}, and 200 \textit{human description} (where definitions are from a human, instead of a dictionary). The evaluation metrics are retrieval accuracies at 1, 10 and 100, as well as the median and standard deviation of the target words' ranks.\footnote{Previous papers might use ``standard deviation'' and ``rank variance'' interchangeably. We stick to ``standard deviation''.}

\textbf{\textsc{Chang}}: definition modelling is experimented on \citet{chang-chen-2019-word}'s data from the Oxford English Dictionary. Each instance is a tuple of a word, its usage (context), and a definition. The data has two splits: \textit{seen} and \textit{unseen}. The \textit{unseen} split we use consists of 530k training instances, and the test set is 1k words paired with 16.0k definitions and context. Performance is measured by corpus-level \textsc{Bleu} from \textsc{Nltk}, and \textsc{Rouge-L} F1\footnote{\url{https://github.com/pltrdy/rouge}} \citep{papineni-etal-2002-bleu, lin-2004-rouge, nltk2009}.

\begin{table*}[tbh]
\centering
\begin{tabular}{|l|cccc|cccc|} 
 \hline
  & \multicolumn{4}{c|}{unseen} & \multicolumn{4}{c|}{human description} \\
 \cline{2-9}
  & \makecell{median\\rank} & \makecell{acc@\\1/10/100} & \makecell{rank\\std.\textsuperscript{\textdagger}} & \makecell{real\\std.} & \makecell{median\\rank} & \makecell{acc@\\1/10/100} & \makecell{rank\\std.\textsuperscript{\textdagger}} & \makecell{real\\std.} \\
 \hline
 OneLook.com  & - & - & - & - & 5.5& .33/.54/.76 & 332 & - \\
 bag-of-words & 248 & .03/.13/.39 & 424 & - & 22 &.13/.41/.69 & 308 & - \\
 RNN          & 171 & .03/.15/.42 & 404 & - & 17 & .14/.40/.73 & 274 & - \\
 category inference & 170 & .05/.19/.43 & 420 & - & 16 & .14/.41/.74 & 306 & - \\
 multi-sense  & 276 & .03/.14/.37 & 426 & - & 1000 &.01/.04/.18 & 404 & - \\
 super-sense  & 465 & .02/.11/.31 & 454 & - & 115 &.03/.15/.47 & 396 & - \\
 multi-channel & 54 & .09/.29/.58 & \textbf{358} & - & \textbf{2} & \textbf{.32}/.64/.88 & 203 & - \\
 \hline
 Transformer & 79 & .01/.14/.59 & 473 & 125 & 27 & .05/.23/.87 & 332 & 49 \\
 unified     & \textbf{18} & \textbf{.13}/\textbf{.39}/\textbf{.81} & {386}  & \textbf{93} & 4 & .22/.64/\textbf{.97} & \textbf{183} & \textbf{30} \\
 \ + share embed  & 20 & .08/.36/.77 & 410 & 99 & 4 & .23/\textbf{.65}/\textbf{.97} & \textbf{183} & 32 \\
 \hline
\end{tabular}
\caption{Reverse dictionary results on the \textsc{Hill} data with past results from \citet{zhang-etal-2020-multi-channel}'s re-run. \textsuperscript{\textdagger}They force-set a word rank larger than 100 to 1000 which affected std.; we follow suit for comparison, and also include the real std.}
\label{tab:1}
\end{table*}

\subsection{The questionable \textit{seen} test set}
Understandably, a dictionary needs to ``memorize'' word entries, so both \textsc{Hill} and \textsc{Chang} supply a \textit{seen} test drawn from training data. However, this is impractical in deep learning, for it implicitly encourages overfitting. Further, the foremost function of a neural dictionary is to deal with unseen words and definitions; otherwise, a traditional rule-based one suffices. We hence omit evaluation on \textit{seen} sets and request future research to not focus on it.

\subsection{System configurations}
Our baselines are 4-layer Transformer blocks: a Transformer encoder for reverse dictionary, and a Transformer decoder for definition modelling. Hyperparameter searches are detailed in Appendix~\ref{sec:appendix-hyper-parameters}. We tokenize training definitions into an open vocabulary by whitespace. We use cross-entropy for definition tokens and mean squared error (MSE) as the embedding distance.

Our proposed model essentially connects and trains the above two baselines with an extra shared layer. The layer has the same size as the input embeddings and a residual connection \citep{he-etal-2016-deep-residual}. As an additional variant, we tie both Transformer blocks' embedding and output layers \citep{press-wolf-2017-using}. This is only possible with our multi-task framework, since a Transformer block baseline does not have both encoder and decoder embeddings. The unified model optimizes roughly twice as many parameters as a single-task baseline; in other words, when performing both tasks, our system is of the same size as the baseline models.

For reverse dictionary, we compare with a number of existing works: {OneLook.com}, bag-of-words, RNN \citep{hill-etal-2016-learning-understand}, category inference \citep{morinaga-yamaguchi-2018-improvement}, multi-sense \citep{kartsaklis-etal-2018-mapping}, super-sense \citep{pilehvar-2019-importance} and multi-channel \cite{zhang-etal-2020-multi-channel}. Following \citet{zhang-etal-2020-multi-channel} we embed target words with 300d \textit{word2vec} \citep{mikolov-etal-2013-efficient}, but definition tokens are encoded into 256d embeddings to train from scratch, instead of pre-trained embeddings.

For definition modelling, words are embedded by 768d \textit{BERT-base-uncased}, while definition token embeddings are initialized randomly. We include RNN \citep{noraset-etal-2017-definition} and xSense \citep{chang-etal-2018-xsense} for reference but not \citet{chang-chen-2019-word}'s results from an oracle retrieval experiment.

Our choice of word embedders aligns with previous works, which ensures that comparison is fair and improvement comes from the model design. It is also worth noting that we train separate models on HILL and CHANG data to evaluate reverse dictionary and definition modelling performances respectively.

\subsection{Results}
\paragraph{Reverse dictionary} results in Table~\ref{tab:1} show a solid baseline, which our proposed models significantly improve upon. Compared to previous works, we obtain the best ranking and accuracies on \textit{unseen} words. On \textit{human descriptions} our models yield compelling accuracies with the best standard deviation, indicating a consistent performance.

One highlight is that our model attains a superior position without linguistic annotations, other than a word embedder which is always used in previous research. Consequently, ours can be concluded as a more generic framework for this task.

\paragraph{Definition modelling} results are reported in Table~\ref{tab:2}. On the \textit{unseen} test, our model with tied embeddings achieves state-of-the-art scores. The model without it has performance similar to the baseline. Admittedly, while \textsc{Rouge-L} scores look reasonable, the single-digit \textsc{Bleu} might hint at the poor quality of the generation. We conduct human evaluation and discuss that later.

\begin{table}[thb]
\centering
\begin{tabular}{|l|c|c|}
\hline
     & \multicolumn{2}{c|}{unseen} \\
\cline{2-3}
    & \textsc{Bleu} & \textsc{Rouge-L} \\
\hline
 RNN       & 1.7           & 15.8 \\
 xSense    & 2.0  & 15.9 \\
\hline
 Transformer  & 2.4 & 17.9\\
 unified       & 2.2 & 18.5 \\
 \ + share embed  & \textbf{3.0} & \textbf{20.2} \\
\hline
\end{tabular}
\caption{Definition modelling results on the \textsc{Chang} data, with past numbers from \citet{chang-chen-2019-word}'s replicate.}
\label{tab:2}
\end{table}

\section{Analysis and Discussions}
\subsection{Shared embeddings and the vocabulary}\label{sec:vocabulary}
For definition modelling, a shared embedding and output layer brings significant improvement to our proposed approach, but in reverse dictionary, our models arrive at desirable results without it. This is reasonable as well-trained embedding and output layers particularly benefit language generation \citep{press-wolf-2017-using}. It further indicates the usefulness of our unified approach whereby all embedding and output layers can be weight-tied, enabled by concurrently training the two Transformer sub-models for the two tasks.

We have used an open vocabulary, which has weaknesses like being oversized and vulnerable to unknown tokens. Therefore, we add a model with a 25k unigram SentencePiece vocabulary \citep{kudo-richardson-2018-sentencepiece} to definition modelling. All other configurations remain the same as the best-performing model. \textsc{Bleu} and \textsc{Rouge-L} drop to 2.5 and 18.7, proving that an open vocabulary is not an issue in our earlier experiments.

\subsection{Human evaluation on definitions}
Supplementary to the automatic evaluation for definition generation, we run reference-less and reference-based human evaluation, on the Transformer baseline and the best-performing unified model. In a \textit{reference-less} evaluation, a human is asked to pick the preferred output after seeing a word, whereas in a \textit{reference-based} setting, a human sees a reference definition instead. Test instances are sampled, and then the models' outputs are presented in a shuffled order. Two annotators in total evaluated 80 test instances for each setting. Table~\ref{tab:human_eval} records the number of times each model is favoured over the other.

Regardless of the evaluation condition, evaluators often regard the unified model's outputs as better. Especially in the reference-less scenario, which resembles a real-life application of definition generation, our unified model wins notably.

\begin{table}[thb]
\centering
\setlength{\tabcolsep}{0.65ex}
\begin{tabular}{|l|c|c|}
\hline
    & reference-less & reference-based \\
\hline
 Transformer      & 25 (31\%) & 32 (40\%) \\
 unified  & \textbf{50 (63\%)} & \textbf{42 (53\%)}  \\
\hline
\end{tabular}
\caption{Chances a model's output is preferred by human evaluators. Columns do not add up to 80 (100\%) because we do not count when both models generated the same output.}
\label{tab:human_eval}
\end{table}

\subsection{Ablation studies on the objectives}
Our models are trained with five losses from five tasks: definition modelling, reverse dictionary, two reconstruction tasks and a shared embedding similarity task. In contrast to the full 5-task model, we try to understand how multiple objectives influence learning, by excluding certain losses.

We first remove reconstruction losses to form a 3-task model that learns reverse dictionary, definition modelling and embedding similarity. This is the minimum set of tasks required to train the full architecture and to ensure words and definitions are mapped to the same representation. Then we designate 1-task models to learn either reverse dictionary or definition modelling depending on the baseline it is compared to. Such a model is deeper than the baseline Transformer but partly untrained. 

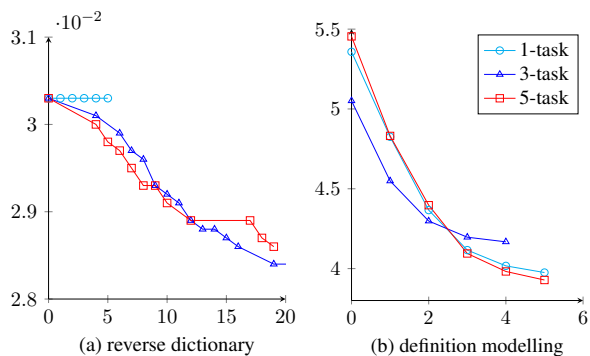
\begin{figure}[ht!]
\centering
\resizebox{0.25\textwidth}{!}{%
\begin{tikzpicture}
\begin{groupplot}[
    group style={
        group name=revdic,
        group size=1 by 2
    },
    width=6cm,
    xmin=0, xmax=20,
    xlabel=(a) reverse dictionary
]

\nextgroupplot[ymin=0.028,ymax=0.031,
               axis y line=left,
               axis x line=bottom,
               height=6.5cm]
\addplot[
    color=cyan,
    mark=o,
    ]
    coordinates {
    (0,0.0303)(1,0.0303)(2,0.0303)(3,0.0303)(4,0.0303)(5,0.0303)
    };
\addplot[
    color=blue,
    mark=triangle,
    ]
    coordinates {
    (0,0.0303)(4,0.0301)(6,0.0299)(7,0.0297)(8,0.0296)(9,0.0293)(10,0.0292)(11,0.0291)(12,0.0289)(13,0.0288)(14,0.0288)(15,0.0287)(16,0.0286)(19,0.0284)(28,0.0284)(33,0.0283)(34,0.0283)(37,0.0282)(45,0.0281)
    };
\addplot[
    color=red,
    mark=square,
    ]
    coordinates {
    (0,0.0303)(4,0.0300)(5,0.0298)(6,0.0297)(7,0.0295)(8,0.0293)(9,0.0293)(10,0.0291)(12,0.0289)(17,0.0289)(18,0.0287)(19,0.0286)
    };

\end{groupplot}
\end{tikzpicture}
}%
\resizebox{0.24\textwidth}{!}{%
\begin{tikzpicture}
\begin{groupplot}[
    group style={
        group name=defmod,
        group size=1 by 2
    },
    width=6cm,
    xmin=0, xmax=6,
    xlabel=(b) definition modelling,
    legend cell align={left}
]

\nextgroupplot[ymin=3.8,ymax=5.5,
               axis y line=left,
               axis x line=bottom,
               height=6.8cm]
\addplot[
    color=cyan,
    mark=o,
    ]
    coordinates {
    (0,5.3579)(1,4.8256)(2,4.3663)(3,4.1158)(4,4.0170)(5,3.9753)
    };
    \addlegendentry{1-task};
\addplot[
    color=blue,
    mark=triangle,
    ]
    coordinates {
    (0,5.0512)(1,4.5491)(2,4.2980)(3,4.1966)(4,4.1676)
    };
    \addlegendentry{3-task};
\addplot[
    color=red,
    mark=square,
    ]
    coordinates {

    (0,5.4544)(1,4.8312)(2,4.3971)(3,4.0946)(4,3.9820)(5,3.9284)
    };
    \addlegendentry{5-task};

\end{groupplot}

\end{tikzpicture}
}%
\vspace{-2ex}\caption{Validation losses (y-axis) against epochs (x-axis).}\vspace{-1ex}
\label{fig:ablation_losses}
\end{figure}

We run the ablation investigation on both reverse dictionary and definition modelling tasks. We log training dynamics in Figure~\ref{fig:ablation_losses}: embedding MSE against epochs for reverse dictionary, and generation cross-entropy against epochs for definition modelling. The curve plotting stops when validation does not improve.

As Figure~\ref{fig:ablation_losses}a shows, the single-task \textsc{Hill} model does not converge, probably because in reverse dictionary the Transformer block is far away from the output end, and only receives small gradients from just one loss. The 3-task and 5-task models display similar losses, but the 3-task loss curve is smoother. In Figure~\ref{fig:ablation_losses}b for definition modelling, the 3-task model trains the fastest, but 1-task and 5-task models reach better convergence. It implies that learning more than one task is always beneficial compared to single-task training; reconstruction is sometimes helpful but not crucial.

\section{Conclusion}
We build a multi-task model for reverse dictionary and definition modelling. The approach records strong numbers on public datasets. Our method delegates disambiguation to BERT and minimizes dependency on linguistically annotated resources, so it can potentially be made cross-lingual and multilingual. A limitation is that the current evaluation centers on English, without exploring low-resource languages, which could be impactful extensions that benefit the community.

\section*{Acknowledgements}
We are grateful to Kenneth Heafield and the reviewers of this paper for their feedback. Pinzhen Chen is funded by the High Performance Language Technologies project with Innovate UK. Zheng Zhao is supported by the UKRI Centre for Doctoral Training in Natural Language Processing (UKRI grant EP/S022481/1).

\bibliography{custom}
\bibliographystyle{acl_natbib}

\appendix
\section{Hyperparameters and Computation}
\label{sec:appendix-hyper-parameters}

Our model configuration search is summarized here. We adjusted the hyperparameters for the baseline using the validation set, and kept the values unchanged for the proposed model which joins two baseline Transformer blocks. We list all hyperparameters in Table~\ref{tab:config}, and highlight the selected ones in bold if multiple values were tried out. The trial is carried out one by one for each hyperparameter. On a single Nvidia GeForce GTX 1080 Ti, it takes 60 hours for a reverse dictionary model to converge; a definition modelling model converges after 6 hours on a single Nvidia GeForce RTX 2080 Ti.

\begin{table}[htb]
\small
\centering
\begin{tabular}{|l|l|} 
\hline
 word embed.             & \textsc{Hill}: word2vec \\
                         & \textsc{Chang}: BERT-base-uncased \\
 word embed. dim.        & \textsc{Hill}: 300 \\
                         & \textsc{Chang}: 768 \\
 definition tokenizer    & whitespace \\
 def. token embed.       & none, trained from one-hot \\
 def. token embed. dim.  & 256 \\
\hline
 training toolkit       & PyTorch \citep{paszke-etal-2019-pytorch} \\
 stopping criterion     & 5 non-improving validations \\
 learning rate          & 1e-3, \textbf{1e-4}, 1e-5 and 1e-6 \\
 optimizer              & Adam \citep{kingma-ba-2015-adam} \\
 beta1, beta2           & 0.9, 0.999 \\
 weight decay           & 1e-6 \\
 embedding loss         & \textbf{MSE}, cosine (failed to converge) \\
 token loss             & cross-entropy \\
 training batch size    & \textsc{Hill}: 256 \\
                        & \textsc{Chang}: 128 \\
 decoding batch size    & 1 \\
 decoding beam size     & \textbf{6}, 64 \\
\hline
 Transformer depth      & \textbf{4}, 6 \\
 Transformer head       & \textbf{4}, 8 \\
 Transformer dropout    & 0.1, \textbf{0.3} \\
 def. token dropout     & \textbf{0}, 0.1 \\ 
 linear layer dropout   & 0.2 \\
 linear layer dim.      & \textsc{Hill}: 256 \\ 
                        & \textsc{Chang}: 768 \\
 shared layer dim.      & \textsc{Hill}: 256 \\ 
                        & \textsc{Chang}: 768 \\
\hline
 trainable parameters   & \textsc{Hill}: 35.1M \\ 
                        & \textsc{Chang}: 62.7M \\
\hline
\end{tabular}
\caption{Model and training configurations.}
\label{tab:config}
\end{table}

\end{document}